\documentclass[runningheads]{llncs}

\usepackage{times}
\usepackage{graphicx}
\usepackage{latexsym}
\usepackage{hyperref}

\usepackage{booktabs}       
\usepackage{amsfonts}       

\usepackage{amsmath}
\usepackage{bbm}
\usepackage{soul} 
\usepackage{graphicx}
\usepackage{subcaption}
\usepackage{mfirstuc} 
\graphicspath{ {./figures/} }
\usepackage{listings}
\newcommand{\Reals}{\ensuremath\mathbb{R}}

\newcommand{\Ex}[2][]{\ensuremath{\mathop\mathbb{E}_{#1}\left[ {#2} \right]}}
\newcommand{\Prob}[2][]{\ensuremath{\mathop\mathsf{Pr}_{#1}\left[ {#2} \right]}}
\newcommand{\Set}[1]{\ensuremath{\left\{ {#1} \right\}}}
\newcommand{\Abs}[1]{\ensuremath{\left| {#1} \right|}}

\newcommand{\ismale}{x_{male}}
\newcommand{\islifter}{x_{lift}}
\newcommand{\fmale}{f_{male}}
\newcommand{\fmalelifter}{f_{both}}

\usepackage{xcolor}

\DeclareMathOperator{\pre}{pre}

\DeclareMathOperator*{\argmin}{arg\,min}

\MFUnocap{a}
\MFUnocap{an}
\MFUnocap{and}
\MFUnocap{in}
\MFUnocap{of}
\MFUnocap{over}
\MFUnocap{the}
\MFUnocap{with}
\MFUnocap{without}
\newcommand\inp{\pmb{x}}
\newcommand\cinp{\pmb{z}}
\newcommand\rinp{\pmb{r}}
\newcommand\randref{\pmb{R}}
\newcommand\randorder{\pmb{O}}

\newcommand\randg{\pmb{G}}
\newcommand\sampleg{\pmb{g}}
\newcommand\distr{\mathcal{D}}
\newcommand\distrinp{\mathcal{D}^{inp}}
\newcommand\distrjm{\mathcal{D}^{J.M.}}
\newcommand\distrunif{\mathcal{U}}

\newcommand\distraccept{\distr^\mathit{accept}}
\newcommand{\puregamef}{v}
\newcommand{\gamef}{v_{\inp}}
\newcommand{\gamefcond}{\gamef^{cond}}
\newcommand{\gamefmarg}{\gamef^{inp}}
\newcommand{\gamefjmarg}{\gamef^{J.M.}}
\newcommand{\gamefunif}{\gamef^{unif}}
\newcommand{\gamefgeneric}{v_{\inp, \distrref}}
\newcommand{\gamefsref}{v_{\inp,\rinp}}
\newcommand{\gamefsrandref}{v_{\inp, \randref}}

\newcommand{\players}{\mathcal{M}}
\newcommand{\features}{\mathcal{M}}
\newcommand{\probphi}{\pmb{\Phi}_{\inp, \randref}}

\newcommand{\distrref}{\mathcal{D}^{ref}}

\begin{document}

\title{The Explanation Game: Explaining Machine Learning Models Using Shapley Values}
%
%
\author{Luke Merrick\inst{1} \and
Ankur Taly\inst{1}}
%
%
\institute{
Fiddler Labs, Palo Alto, USA\\
\email{\{luke,ankur\}@fiddler.ai}}
\maketitle              

\begin{abstract}
A number of techniques have been proposed to explain a machine learning model's prediction by attributing it to the corresponding input features. Popular among these are techniques that apply the Shapley value method from cooperative game theory. While existing papers focus on the axiomatic motivation of Shapley values, and efficient techniques for computing them, they offer little justification for the game formulations used, and do not address the uncertainty implicit in their methods' outputs. For instance, the popular SHAP algorithm's formulation may give substantial attributions to features that play no role in the model.
In this work, we illustrate how subtle differences in the underlying game formulations of existing methods can cause large differences in the attributions for a prediction. We then present a general game formulation that unifies existing methods, and enables straightforward confidence intervals on their attributions. Furthermore, it allows us to interpret the attributions as \emph{contrastive explanations} of an input relative to a distribution of reference inputs.
We tie this idea to classic research in cognitive psychology on contrastive explanations, and propose a conceptual framework for generating and interpreting explanations for ML models, called \emph{formulate, approximate, explain} (FAE).
 We apply this framework to explain black-box models trained on two UCI datasets and a Lending Club dataset.
\end{abstract}

\section{\uppercase{Introduction}}\label{attributions}
Complex machine learning models are rapidly spreading to high stakes tasks such as credit scoring, underwriting, medical diagnosis, and crime prediction. Consequently, it is becoming increasingly important to interpret and explain individual model predictions to decision-makers, end-users, and regulators. A common form of model explanations are based on \emph{feature attributions}, wherein a score (\emph{attribution}) is ascribed to each feature in proportion to the feature's contribution to the prediction. Over the last few years there has been a surge in feature attribution methods, with methods based on Shapley values from cooperative game theory being prominent among them \cite{vstrumbelj2014explaining,datta2016algorithmic,lundberg2017unified,shapley_mahalanobis,lundberg2018consistent,shapley_polynomial,chen2018shapley}. 

Shapley values~\cite{shapley1953value} provide a mathematically fair and unique method to attribute the payoff of a cooperative game to the players of the game. 
Recently, there have been a number of Shapley-value-based methods for attributing an ML model's prediction to input features. 
Prominent among them are SHAP and KernelSHAP \cite{lundberg2017unified}, TreeSHAP~\cite{lundberg2018consistent}, QII~\cite{datta2016algorithmic}, and IME~\cite{vstrumbelj2010efficient}.
In applying the Shapley values method to ML models, the key step is to setup a cooperative game whose players are the input features and whose payoff is the model prediction.
Due to its strong axiomatic guarantees, the Shapley values method is emerging as the de facto approach to feature attribution, and some researchers even speculate that it may be the only method compliant with legal regulation such as the General Data Protection Regulation's ``right to an explanation''~\cite{shapley_mahalanobis}.

In this work, we study several Shapley-value-based explanation techniques. 
Paradoxically, while all  techniques lay claim to the axiomatic uniqueness of Shapley values, we discover that they yield significantly different attributions for the same input even when evaluated exactly (without approximation).
In some cases, we find the attributions to be completely counter-intuitive.
For instance, in Section~\ref{sec:motivating_example}, we show a simple model for which the popular SHAP method gives substantial attribution to a feature that is irrelevant to the model function.
We trace this shortcoming and the differences across existing methods to the varying cooperative games formulated by the methods.\footnote{We note that this shortcoming, and the multiplicity of game formulations has also been noted in parallel work~\cite{mukund2019shapley,janzing2019feature}} We refer to such games as \emph{explanation games}.
Unfortunately, while existing methods focus on the axiomatic 
motivations of Shapley values, they offer little justification for the design choices made in their explanation game formulations. The goal of this work is to shed light on these design choices, and their implications on the resulting attributions.

Our main technical result shows that various existing techniques can be unified under a common game formulation parameteric on a reference distribution.
The Shapley values of this unified game formulation can be decomposed into the Shapley values of \emph{single-reference games} that model a feature's absence by replacing its value with the corresponding value from a specific \textit{reference input}.

This decomposition is beneficial in two ways. First, it allows us to efficiently compute confidence intervals and other supplementary information about attributions, a notable advancement over existing methods (which lack confidence intervals even though they approximate metrics of random variables using finite samples). Second, it offers conceptual clarity. It unlocks the interpretation that attributions explain the prediction at an input \emph{in contrast} to other reference inputs.
The attributions vary across existing methods as each method chooses a different reference distribution to contrast with. 
We tie the idea to classic research in cognitive psychology, and  propose a conceptual \emph{formulate, approximate, explain} (FAE) framework to create Shapley-value-based \emph{contrastive} feature attributions.
The goal of the framework is to produce attributions that are not only axiomatically justified, but also relevant to the underlying explanation question.


We illustrate our ideas via case studies on models trained on two UCI datasets (Bike Sharing and Adult Income) and a Lending Club dataset. We find that in these real-world situations, explanations generated using our FAE framework uncover important patterns that previous attribution methods cannot identify.
In summary, we make the following key contributions:
\begin{itemize}
    \item We highlight several shortcomings of existing Shapley-value-based feature attribution methods (Sections~\ref{sec:motivating_example}), and analyze the root cause of these issues (Section~\ref{subsec:existing_games}).
    \item We present a novel game formulation that unifies existing methods (Section~\ref{subsec:reformulation}), and helps characterize their uncertainty with confidence intervals (Section~\ref{sec:ci_theory}).
    \item We offer a novel framework for creating and interpreting attributions (Section~\ref{subsec:fae}), and demonstrate its use through case studies (Section~\ref{sec:case_studies}).
\end{itemize}

\section{A motivating example}\label{sec:motivating_example}
\begin{table}[!t]
    \small
    \centering
    \begin{tabular}{lllll}
    \toprule
    $\ismale$ & $\islifter$ & $\Prob{\pmb{X} = \inp}$ & $\fmale(\inp)$ & $\fmalelifter(\inp)$  \\
    \midrule
    0                 & 0                               & 0.1    & 0.0                         & 0.0        \\
    0                 & 1                               & 0.0    & 0.0                         & 0.0        \\
    1                 & 0                               & 0.4    & 1.0                         & 0.0        \\
    1                 & 1                               & 0.5    & 1.0                         & 1.0        \\
    \bottomrule
    \end{tabular}
    \caption{Input distribution and model outputs for the mover hiring system example.}    \label{table:mover_probability}
\end{table}

\begin{table*}[!t]
\small
  \centering
  \begin{tabular}{lll|ll}
    \toprule
                                            & \multicolumn{2}{c}{$\fmale$}                                    & \multicolumn{2}{c}{$\fmalelifter$} \\
    \textbf{Payoff formulation}                    & $\phi_1$ (\textit{male}) & $\phi_2$ (\textit{lifting}) 
    & $\phi_1$ (\textit{male}) & $\phi_2$ (\textit{lifting}) \\
    \midrule
    SHAP                     & 0.05                   & 0.05             & 0.028                   & 0.472   \\
    KernelSHAP                    & 0.10                   & 0.00             & 0.050                   & 0.450   \\
    QII                     & 0.10                   & 0.00            & 0.075                   & 0.475   \\
    IME                     & 0.50                   & 0.00             & 0.375                   & 0.375   \\
    \bottomrule
  \end{tabular}
  \caption{Attributions for the input $\ismale = 1, \islifter  = 1$.}
  \label{table:two_variable_expected}
\end{table*}

To probe existing Shapley-value-based model explanation methods, we evaluate them on two toy models for which it is easy to intuit correct attributions.
We leverage a modified version of the example provided
in \cite{datta2016algorithmic}: a system that recommends whether a moving company should hire a mover applicant.
The input vector to both models comprises two binary features ``is male'' and ``is good lifter'' (denoted by $\pmb{x} = (\ismale, \islifter)$), and output a recommendation score between 0 (``no hire'') and 1 (``hire'').
We define two models --- $\fmale(\pmb{x}) ::= \ismale$ (only hire males), and $\fmalelifter(\pmb{x}) ::= \ismale \land \islifter$ (only hire males who are good lifters). Table \ref{table:mover_probability} specifies a probability distribution over the input space, along with the predictions from the two models. 

Consider the input $\pmb{x} = (1,1)$ (i.e. a male who is a good lifter), for which both models output a recommendation score of 1. Table~\ref{table:two_variable_expected} lists the attributions from several existing methods. Focusing on the relative attribution between $\ismale$ and $\islifter$, we make the following surprising observations. First, even though $\islifter$ is irrelevant to $\fmale$, the SHAP algorithm\footnote{As defined by Equation 9 in \cite{lundberg2017unified}.} results in equal attribution to both features. This contradicts our intuition around the ``Dummy'' axiom of Shapley values, which states that attribution to a player (feature) that never contributes to the payoff (prediction) must be zero.

Additionally, the SHAP attributions present a misleading picture from a fairness perspective: $\fmale$ relies solely on $\ismale$, yet the attributions do not reflect this bias and instead claim that the model uses both features equally.
Second, although $\fmalelifter$ treats its features symmetrically, and $\inp$ has identical values in both its features, many of the methods considered do not provide symmetrical attributions. 
This again is intuitively at odds with the ``Symmetry'' axiom of Shapley values, which states that players (features) that always contribute equally to the payoff (prediction) must receive equal attribution.
These unintuitive behaviors surfaced by the above observations demand an in-depth study of the internal design choices of these methods. We carry out this study in Section~\ref{sec:explanation_games}.

\section{\uppercase{Preliminaries}}\label{sec:preliminaries}

\subsection{Additive feature attributions}\label{subsec:attributions}
\emph{Additive feature attributions} \cite{lundberg2017unified} are attributions that sum to the difference between the explained model output $f(\inp)$ and a reference output value $\phi_0$. In practice, $\phi_0$ is typically an average model output or model output for a domain-specific ``baseline'' input (e.g. an empty string for text sentiment classification).

\begin{definition}[Additive feature attributions]\label{def:attributions}
Suppose $f: \mathcal{X} \rightarrow \Reals$ is a model mapping an $M$-dimensional feature space $\mathcal{X}$ to real-valued predictions.
Additive feature attributions for $f(\inp)$ at input $\inp = (x_1, \ldots, x_M) \in \mathcal{X}$ comprise of a reference (or baseline) attribution $\phi_0$ and feature attributions $\pmb{\phi} = (\phi_1, \phi_2, \ldots, \phi_M)$ corresponding to the $M$ features such that $f(\inp) = \phi_0 + \sum_{i=1}^M \phi_i$.
\end{definition}

There currently exist a number of competing methodologies for computing these attributions (see~\cite{ancona2018gradients}).
Given the difficulty of empirically evaluating attributions, several methods offer an axiomatic justification, often through the Shapley values method.

\subsection{Shapley values}\label{subsec:shapley_values}
The Shapley values method is a classic technique from game theory that fairly attributes the total payoff from a cooperative game to the game's players \cite{shapley1953value}. Recently, this method has found numerous applications in explaining ML models (e.g. \cite{cohen2005feature,lundberg2017unified,data_shapley}).

Formally, a cooperative game is played by a set of players  $\players = \{1, \ldots, M\}$ termed the \emph{grand coalition}. The game is characterized by a set
function $\puregamef:2^{\players} \rightarrow \Reals$ such that $\puregamef(S)$ is the payoff for any coalition of players 
$S \subseteq \players$, and $\puregamef(\emptyset) = 0$.
Shapley values are built by examining the marginal contribution of a player to an existing coalition $S$, i.e., $\puregamef(S \cup \Set{i}) - \puregamef(S)$. The Shapley value of a player $i$, denoted $\phi_i(\puregamef)$, is a certain weighted aggregation of its marginal contribution to all possible coalitions of players.
\small
\begin{equation}\label{eq: shapley}
    \phi_i(\puregamef) = \frac{1}{M} \sum_{S \subseteq \players \setminus \Set{i}} \binom{M - 1}{\Abs{S}}^{-1} \left(\puregamef(S \cup \Set{i}) - \puregamef(S)\right)
\end{equation}
\normalsize
The Shapley value method is the unique method satisfying four desirable axioms: \emph{Dummy}, \emph{Symmetry}, \emph{Efficiency}, and \emph{Linearity}. We informally describe the axioms in Appendix~\ref{appendix:shapley_axioms}, and refer the reader to~\cite{young85} for formal definitions and proofs.

\subsubsection{Approximating Shapley values}\label{subsec:approx_sv}
Computing Shapley values involves evaluating the game payoff for every possible coalition of players. This makes the computation exponential in the number of players. For games with few players, it is possible to exactly compute the Shapley values, but for games with many players, the Shapley values can only be approximated. Recently there has been much progress towards the efficient approximation of Shapley values. In this work we focus on a simple sampling approximation, presenting two more popular techniques in the Appendix~\ref{appendix:shapley_approx}. We refer the reader to \cite{maleki2013bounding,chen2018shapley,shapley_mahalanobis,hunt2019ai,shapley_polynomial} for a fuller picture of recent advances in Shapley value approximation.

A simple sampling approximation (used by~\cite{data_shapley}, among other works) relies on the fact that the Shapley value can be expressed as the expected marginal contribution a player has when players are added to a coalition in a random order.
Let $\pi(M)$ be the ordered set of permutations of $M$, and $\randorder$ be an ordering randomly sampled from $\pi(M)$. Let $\pre_i(\randorder)$ be the set of players that precede player $i$ in $\randorder$. The Shapley value of player $i$ is the expected marginal contribution of the player under all possible orderings of players.
\small
\begin{equation}\label{eq: perm_sample}
    \phi_i(\puregamef) = \Ex[\randorder\sim\pi(M)]{ \puregamef(\pre_i(\randorder) \cup \Set{i}) - \puregamef(\pre_i(\randorder))}
\end{equation}
\normalsize
By sampling a number of permutations and averaging the marginal contributions of each player, we can estimate this expected value for each player and approximate each player's Shapley value.

\section{\uppercase{Explanation Games}}\label{sec:explanation_games}
In order to explain a model prediction with the Shapley values method, it is necessary to formulate a cooperative game with players that correspond to the features and a payoff that corresponds to the prediction. In this section, we analyze the methods examined in Section~\ref{sec:motivating_example}, and show that their surprising attributions are an artifact of their game formulations.
We then discuss a unified game formulation and its decomposition to \emph{single-reference games}, enabling conceptual clarity about the meanings of existing methods' attributions.

\subsubsection{Notation}
Let $\distrinp$ be the input distribution, which characterizes the process that generates model inputs.
We denote the input of an explained prediction as $\pmb{x} = (x_1, \ldots, x_M)$ and use $\rinp$ to denote another ``reference'' input.
We use boldface to indicate when a variable or function is vector-valued, and 
capital letters for random variable inputs
(although $S$ continues to represent the set of contributing players/features).
Thus, $x_i$ is a scalar input, $\inp$ is an input vector, and $\pmb{X}$ is a \emph{random} input vector.
We use $\inp_S = \Set{x_i : i \in S}$ to represent a sub-vector of features indexed by $S$. This notation is also extended to random input vectors $\pmb{X}$. 
Lastly, we introduce the \emph{composite input} $\cinp(\inp, \rinp, S)$, which
agrees with the input $\inp$ on all features in $S$ and with $\rinp$ on all features not in $S$.
Note that $\cinp(\inp, \rinp, \emptyset) = \rinp$, and $\cinp(\inp, \rinp, \features) = \inp$.
\begin{equation}\label{eq: cinp}
\small
\cinp(\inp, \rinp, S) = (z_1, z_2, ..., z_M) \text{, where }  z_i =     \begin{cases} 
                                                                                    x_i & i \in S \\
                                                                                    r_i & i \notin S
                                                                                \end{cases}
\end{equation}
\normalsize

\subsection{Existing game formulations}\label{subsec:existing_games}
The explanation game payoff function $\gamef$ must be defined for every feature subset $S$ such that $\gamef(S)$ captures the contribution of $\inp_S$ to the model's prediction. This allows us to compute each feature's possible marginal contributions to the prediction and derive its Shapley value (see Section~\ref{subsec:shapley_values}).

By the definition of \textit{additive feature attributions} (Definition~\ref{def:attributions}) and the Shapley values' Efficiency axiom, we must define $\gamef(\features) ::= f(\inp) - \phi_0$ (i.e. the payoff of the full coalition must be the difference between the explained model prediction and a baseline prediction). Although this definition is fixed, it leaves us the challenge of coming up with the payoff when some features do not contribute (that is, when they are \emph{absent}).

We find that all existing approaches handle this feature-absent payoff by randomly sampling absent features according to a particular  \emph{reference} distribution and then computing the expected value of the prediction. The resulting game formulations differ from one another only in the reference distribution they use. Additionally, we note that in practice small samples are used to approximate the expected value present in these payoff functions. This introduces a significant source of attribution uncertainty not clearly quantified by existing work.

\subsubsection{Conditional distribution}
The game formulation of SHAP~\cite{lundberg2017unified}, TreeSHAP~\cite{lundberg2018consistent}, and \cite{shapley_mahalanobis} simulates feature absence by sampling absent features from the conditional distribution based on the values of the present (or contributing) features:
\begin{equation}\label{game:cond}
\small
\gamefcond(S) = 
    \Ex[\randref \sim \distrinp]{f(\cinp(\inp, \randref, S))~|~\randref_S = \inp_S}
    - \Ex[\randref \sim \distrinp]{f(\randref)}
\end{equation}
Unfortunately, this is not a proper simulation of feature absence as it does not break correlations between features~\cite{janzing2019feature}. This could lead to unintuitive attributions.
For instance, in the $\fmale$ example from Section~\ref{sec:motivating_example}, it causes the the irrelevant feature $\islifter$ to receive a nonzero attribution.
Specifically, since the event $\ismale=1$ is correlated\footnote{In this context, \emph{correlation} refers to general statistical dependence, not just a nonzero Pearson correlation coefficient.} with $\islifter = 1$, once $\islifter = 1$ is given, the expected prediction becomes 1. This causes the $\islifter$ feature to have a non-zero marginal contribution (relative to when both features are absent), and therefore a nonzero Shapley value.
More generally, whenever a feature is correlated with a model's prediction on inputs drawn from $\distrinp$, this game formulation results in non-zero attribution to the feature regardless of whether the feature directly impacts the prediction.

\subsubsection{Input distribution}
Another option for simulating feature absence, which is used by KernelSHAP, is to sample absent features from the corresponding marginal distribution in $\distrinp$:
\begin{equation}\label{game:marg}
\gamefmarg(S) = 
    \Ex[\randref \sim \distrinp]{f(\cinp(\inp, \randref, S))} - \Ex[\randref \sim \distrinp]{f(\randref)}
\end{equation}
Since this formulation breaks correlation with the contributing features, it ensures irrelevant features receive no attribution (e.g. no attribution to $\islifter$ when explaining $\fmale(1, 1) = 1$). We formally describe this property via the \emph{Insentivity} axiom in Section~\ref{subsec:reformulation}.

We note that this formulation is still subject to artifacts of the input distribution, as evident from the asymmetrical attributions when explaining the prediction $\fmalelifter(1,1) = 1$ (see Table~\ref{table:two_variable_expected}). The features receive different attributions because they have different marginal distributions in $\distrinp$, not because they impact the model differently.

\subsubsection{Joint-marginal distribution}
QII~\cite{datta2016algorithmic} simulates feature absence by sampling absent features one at a time from their own univariate marginal distributions. In addition to breaking correlation with the contributing features, this breaks correlation between absent features as well. Formally, the QII formulation uses a distribution we term the ``joint-marginal'' distribution ($\distrjm$), where:
\begin{equation*}
    \Prob[X~\sim~\distrjm]{X=(x_1, \ldots, x_M)} = \prod_{i=1}^{M}\Prob[X\sim\distrinp]{X_i = x_i}
\end{equation*}
The joint-marginal formulation $\gamefjmarg$ is similar to $\gamefmarg$, except that the reference
distribution is $\distrjm$ instead of $\distrinp$:
\begin{equation}\label{game:joint-marg}
\gamefjmarg(S) = 
    \Ex[\randref \sim \distrjm]{f(\cinp(\inp, \randref, S))} - \Ex[\randref \sim \distrjm]{f(\randref)}
\end{equation}
Unfortunately, like $\gamefmarg$, this game formulation is also tied to the input distribution and under-attributes features that take on common values in the background data. This is evident from the attributions for the $\fmalelifter$ model shown in Table~\ref{table:two_variable_expected}.

\subsubsection{Uniform distribution}
The last formulation we study from the prior art simulates feature absence by drawing values from a uniform distribution $\distrunif$ over the entire input space, as in IME~\cite{vstrumbelj2010efficient}.\footnote{It is somewhat unclear whether IME proposes  $\distrunif$ or $\distrinp$, as \cite{vstrumbelj2010efficient} assumes $\distrinp = \distrunif$, while \cite{vstrumbelj2014explaining} calls for values to be sampled from $\mathcal{X}$ ``at random.''}
Completely ignoring the input distribution, this payoff $\gamefunif$ considers all possible feature values (edge-cases and common cases) with equal weighting.
\begin{equation}\label{game:unif}
\gamefunif(S) = 
    \Ex[\randref \sim \distrunif]{f(\cinp(\inp, \randref, S))} - \Ex[\randref \sim \distrunif]{f(\randref)}
\end{equation}
In Table~\ref{table:two_variable_expected}, we see that this formulation yields intuitively correct attributions for $\fmale$ and $\fmalelifter$.
However, the uniform distribution can sample so heavily from irrelevant outlier regions of $\mathcal{X}$ that relevant patterns of model behavior become masked (we study the importance of \emph{relevant references} both theoretically in Section~\ref{subsec:fae} and empirically in Section~\ref{sec:case_studies}).

\subsection{A unified formulation}\label{subsec:reformulation}
We observe that the existing game formulations $\gamefmarg$, $\gamefjmarg$,
and $\gamefunif$ can be unified as a single game formulation $\gamefgeneric$ that
is parameterized by a reference distribution $\distrref$.
\begin{equation}\label{game:unified}
\gamefgeneric(S) = 
    \Ex[\randref \sim \distrref]{f(\cinp(\inp, \randref, S))} - \Ex[\randref \sim \distrref]{f(\randref)}
\end{equation}
For instance, the formulation for KernelSHAP is recovered when $\distrref = \distrinp$, 
and QII is recovered when $\distrref = \distrjm$. In the rest of this section, we discuss
several properties of this general formulation that help us better understand its attributions.
Notably, the formulation $\gamefcond$ cannot be expressed
in this framework; we discuss the reason for this later in this section.

\subsubsection{A decomposition in terms of single-reference games}
We now introduce \emph{single-reference games}, a conceptual building block that helps us interpret the Shapley values of the $\gamefgeneric$ game. A single-reference game $\gamefsref$ simulates feature absence by replacing the feature value with the value from a specific reference input $\rinp$:
\begin{equation}\label{game:single-ref}
\gamefsref(S) = f(\cinp(\inp, \rinp, S)) - f(\rinp)
\end{equation}
The attributions from a single-reference game explain the difference between the prediction for the input and the prediction for the reference (i.e. $\sum_i \phi_i(\gamefsref) = \gamefsref(\features) = f(\inp) - f(\rinp)$, and $\phi_0 = f(\rinp)$).
Computing attributions relative to a single reference point (also referred to as a ``baseline'') is common to several others methods \cite{sundararajan2017axiomatic,shrikumar2017learning,ibm_cem_tabular,shapley_polynomial}. However, while those works seek a neutral ``informationless'' reference (e.g. an all-black image for image models), we find it beneficial to consider arbitrary references and interpret the resulting attributions relative to the reference.
We develop this idea further in our FAE framework (see Section~\ref{subsec:fae}).

We now state Proposition~\ref{prop:reformulation}, which shows how the Shapley values of $\gamefgeneric$ can be 
expressed as the expected Shapley values of a (randomized) single-reference game $\gamefsrandref$, where $\randref \sim \distr$. The proof (provided in Appendix~\ref{appendix:proofs}) follows from the Shapley values' Linearity axiom and the linearity of expectation. 
\begin{proposition}\label{prop:reformulation}
$\pmb{\phi}(\gamefgeneric) = \Ex[\randref \sim \distrref]{\pmb{\phi}(\gamefsrandref)}$
\end{proposition}

Proposition~\ref{prop:reformulation} brings conceptual clarity and practical improvements (confidence intervals and supplementary metrics) to existing methods. It shows that the attributions from existing games ($\gamefmarg$, $\gamefjmarg$, and $\gamefunif$) are in fact differently weighted aggregations of attributions from a space of single-reference games. For instance, $\gamefunif$ weighs attributions relative to all reference points equally, while $\gamefmarg$ weighs them using the input distribution $\distrinp$.

\subsubsection{Insensitivity axiom}\label{sec: insensitivity}
We show that attributions from the game $\gamefgeneric$ satisfy the \emph{Insensitivity} axiom from~\cite{sundararajan2017axiomatic}, which states that a feature that is mathematically irrelevant to the model must receive zero attribution. Formally, a feature $i$ is irrelevant to a model $f$ if for any input, changing the feature does not change the model output. That is, $\forall \inp, \rinp \in \mathcal{X}: \inp_{\features\setminus\{i\}} = \rinp_{\features\setminus\{i\}} \implies f(\inp) = f(\rinp)$.
\begin{proposition}\label{prop:model_dummy}
If a feature $i$ is irrelevant to a model  $f$ then $\phi_i(\gamefgeneric) = 0$ for all distributions $\distrref$.
\end{proposition}
Notably, the $\gamefcond$ formulation does not obey the Insensitivity axiom (a counter-example being the $\fmale$ attributions from Section~\ref{sec:motivating_example}). Accordingly, our general formulation (Equation~\ref{game:unif}) cannot express this formulation. In the rest of the paper, we focus on game formulations that satisfy the Insensitivity axiom. We refer to~\cite{mukund2019shapley} for a comprehensive analysis of the axiomatic guarantees of various game formulations.

\subsection{Confidence intervals on attributions}\label{sec:ci_theory}
Existing game formulations involve computing an expected value (over a reference distribution) in every invocation of the payoff function. In practice, this expectation is approximated via sampling, which introduces uncertainty. The original formulations of these games do not lend themselves well to quantify such uncertainty. We show that by leveraging our unified game formulation, one can efficiently quantify the uncertainty using confidence intervals (CIs).

Our decomposition in Proposition~\ref{prop:reformulation} shows that the attributions themselves can be expressed as an expectation over (deterministic) Shapley value attributions from a distribution of single-reference games. Consequently, we can quantify attribution uncertainty by estimating the standard error of the mean (SEM) across a sample of Shapley values from single-reference games. In terms of the sample standard deviation (SSD), 95\% CIs on the mean attribution ($\bar{\pmb{\phi}}$) from a sample of size $N$ are given by 
\begin{equation}
    \bar{\pmb{\phi}} \pm \frac{1.96 \times \mathsf{SSD}(\Set{\pmb{\phi}(\puregamef_{\inp,\rinp_i})}_{i=1}^{N})}{\sqrt{N}}
\end{equation}
We note that while one could use bootstrap~\cite{efron1986} to obtain CIs, the SEM approach is more efficient as it requires no additional Shapley value computations.

\subsubsection{A unified CI}
As discussed in Section~\ref{subsec:shapley_values}, often the large number of features (players) in an explanation game necessitates the approximation of Shapley values. The approximation may involve random sampling, which incurs its own uncertainty. In what follows, we derive a general SEM-based CI that quantifies the combined uncertainty from sampling-based approximations of Shapley values and the sampling of references.

Let us consider a generic estimator $\hat{\phi}_{i}^{(\randg)}(\puregamef_{\inp,\rinp})$ parameterized by some random sample $\randg$. An example of such an approach is the feature ordering based approximation of Equation~\ref{eq: perm_sample}, for which $\randg = (\randorder_j)_{j=1}^{k}$ represents a random sample of feature orderings, and:
\begin{equation*}
\hat{\phi}_{i}^{(\randg)}(\puregamef_{\inp,\rinp}) = \frac{1}{k} \sum_{j=1}^{k} \puregamef(\pre_i(\randorder_j) \cup \Set{i}) - \puregamef(\pre_i(\randorder_j))
\end{equation*}
As long as the generic $\hat{\phi}_{i}^{(\randg)}$ is an unbiased estimator (like the feature ordering estimator of Equation~\ref{eq: perm_sample}), and $\randg$ and $\randref \sim \distrref$ are sampled independently from one another, we can derive a unified CI using the SEM. By the estimator's unbiasedness and Proposition~\ref{prop:reformulation}, the Shapley value attributions can be expressed as:
\begin{equation}
    \phi_i(\puregamef_{\inp,\distrref}) = \mathop\mathbb{E}_{\randref}\Ex[\randg]{\hat{\phi}_{i}^{(\randg)}(\puregamef_{\inp,\randref})}
\end{equation}
Since $\randg$ is independent of $\randref$, this expectation can be Monte Carlo estimated using the sample mean of the sequence $\left(\hat{\phi}_{i}^{(\sampleg_j)}(\puregamef_{\inp,\rinp_j})\right)_{j=1}^{N}$ (where $\left(\sampleg_j, \rinp_j\right)_{j=1}^{N}$ is a joint sample of $(\randg, \randref)$).
As the attribution recovered by this estimation is simply the mean of a sample from a random variable, its uncertainty can be quantified by estimating the SEM. In terms of the sample standard deviation, 95\% CIs on the mean attribution ($\bar{\pmb{\phi}}$) from a sample of size $N$ are given by:
\begin{equation}
\small
    \bar{\pmb{\phi}} \pm \frac{1.96 \times \mathsf{SSD}\left(\left(\hat{\phi}_{i}^{(\sampleg_j)}(\puregamef_{\inp,\rinp_j})\right)_{j=1}^{N}\right)}{\sqrt{N}}
\end{equation}
\subsection{Formulate, Approximate, Explain}\label{subsec:fae}
So far we studied the explanation game formulations used by existing methods, and noted how the formulations impact the resulting Shapley value attributions.
We show that the attributions explain a prediction \emph{in contrast} to a distribution of references; see Proposition~\ref{prop:reformulation}. 
Existing methods differ in the attribution they produce because each of them picks a different reference distribution to contrast with. 
We also proposed a mechanism to quantify the approximation uncertainty incurred in computing attributions.

We now put these ideas together in a single conceptual framework \emph{formulate, approximate, explain} (FAE). 
Our key insight is that rather than viewing the reference distribution as an implementation detail of the explanation method,
it must by made a first-class argument to the framework. That is, the references must be consciously chosen by the explainee to obtain a specific \emph{contrastive explanation}.

Our emphasis on treating attributions as contrastive explanations stems from cognitive psychology.
Several works in cognitive psychology argue that humans frame explanations of surprising outcomes by contrasting them with to one or more normal outcomes~\cite{kahneman1986norm,miller2017explanation,mittelstadt2019explaining,hesslow88,Hitchcock2009,lipton_1990,holzinger_kandinsky_2019}. 
In our setting, the normal outcomes are the reference predictions that the input prediction is contrasted with.
The attributions essentially explain what drives the prediction at hand away from the reference predictions.
The choice of references may depend on the context of the question, and may vary across explainers and explainees~\cite{kahneman1986norm}.
Moreover, it is important for the references to be \emph{relevant} to the input at hand~\cite{Hitchcock2009}. For instance, if we are explaining why an auto-grading software assigns a B+ to a student's submission, it would be proper to contrast with the submissions that were graded as A- (next higher grade after B+), instead of contrasting with the entire pool of submissions.


\subsubsection{Formulate}
The mandate of the Formulate step is to \textit{generate a contrastive question that specifies one or more relevant references}. The question pins down the distribution $\distrref$ of the chosen references.
For instance, in the grading example above, the references would be all submissions obtaining an A- grade.

\subsubsection{Approximate}
Once a meaningful contrastive question and its corresponding reference distribution $\distrref$ has been formulated,
we consider the distribution of single-reference games whose references are drawn from $\distrref$, and approximate the Shapley values of these games. 
Formally, we approximate the distribution of the random-valued attribution vector $\probphi = \pmb{\phi}(\gamefsrandref)$, where $\randref~\sim~\distrref$.
This involves two steps: (1) sampling a sequence of references $\left( \rinp_i\right)_{i=1}^{N}$ from $\randref \sim \distrref$, and (2) approximating the Shapley value of the single-reference games relative each to reference in $\left( \rinp_i\right)_{i=1}^{N}$. This yields a sequence of approximated Shapley values. It is important to account for the uncertainty resulting from sampling in steps (1) and (2), and quantify it in the Explain step.

\subsubsection{Explain}
In the final step, we must summarize the sampled Shapley value vectors (drawn from $\probphi$) obtained from the Approximate step. One simple summarization would be the presentation of a few representative samples, in the style of the SP-LIME algorithm \cite{ribeiro2016should}. Another simple summarization is the sample mean, which approximates $\Ex{\probphi}$, and is equivalent to the attributions from the unified explanation game $v_{\inp, \distrref}$. 
This is the summarization used by existing Shapley-value-based explanation methods.
When using the sample mean, the framework of Section~\ref{sec:ci_theory} can be used to quantify the uncertainty from sampling.
In addition, one must be careful that the mean does not hide important information. For instance, a feature's attributions may have opposite signs relative to different references. Averaging these attributions will cause them to cancel each other out, yielding a small mean that incorrectly suggests that the feature is unimportant. 
We discuss a concrete example of this in Section~\ref{subsec:case_study_shortcomings}.
At the very least, we recommend confirming through visualization and summary statistics like variance and interquartile range that the mean is a good summarization, before relying upon it.
We discuss a clustering based summarization method in Section~\ref{sec:case_studies} while leaving further research on faithful summarization methods to future work. 
\section{\uppercase{Case studies}}\label{sec:case_studies}
In this section 
we apply the FAE framework to LightGBM \cite{ke2017lightgbm} Gradient Boosted Decision Trees (GBDT) models trained on real data: the UCI Bike Sharing and Adult Income datasets, and a Lending Club dataset.\footnote{In Bike Sharing we model hourly bike rentals from temporal and weather features, in Adult Income we model whether an adult earns more than \$50,000 annually, and in Lending Club we model whether a borrower will default on a loan.} For parsimony, we analyze models that use only five features; complete model details are provided in Appendix~\ref{appendix:model_details}. 
For the Bike Sharing model, we explain a randomly selected prediction of 210 rentals for a certain hour.
For the Adult Income model, we explain a counter-intuitively low prediction for an individual with high \textit{education-num}.
For the Lending Club model, we explain a counter-intuitive rejection (assuming a threshold that accepts  15\% of loan applications) for a high-income borrower. 
In the rest of this section, we present a selection of the results across all three models, while the full set of results are provided in Appendix~\ref{appendix:case_studies}.

\subsection{Shortcomings of existing methods}\label{subsec:case_study_shortcomings}
Recall from Section~\ref{subsec:reformulation} that the attributions from existing methods amount to computing the mean attribution for a distribution of single-reference games $\gamefsrandref$, where the reference $\randref$ is sampled from a certain distribution.
The choice of distribution varies across the methods, which in turns leads to very different attribution. This is illustrated in Table~\ref{table:bikeshare_attributions}
for the Bike sharing model.

\subsubsection{Misleading means}\label{sec:case_study_misleading_means}
\begin{figure}[!t]
  \small
  \centering
  \includegraphics[width=.8\columnwidth]{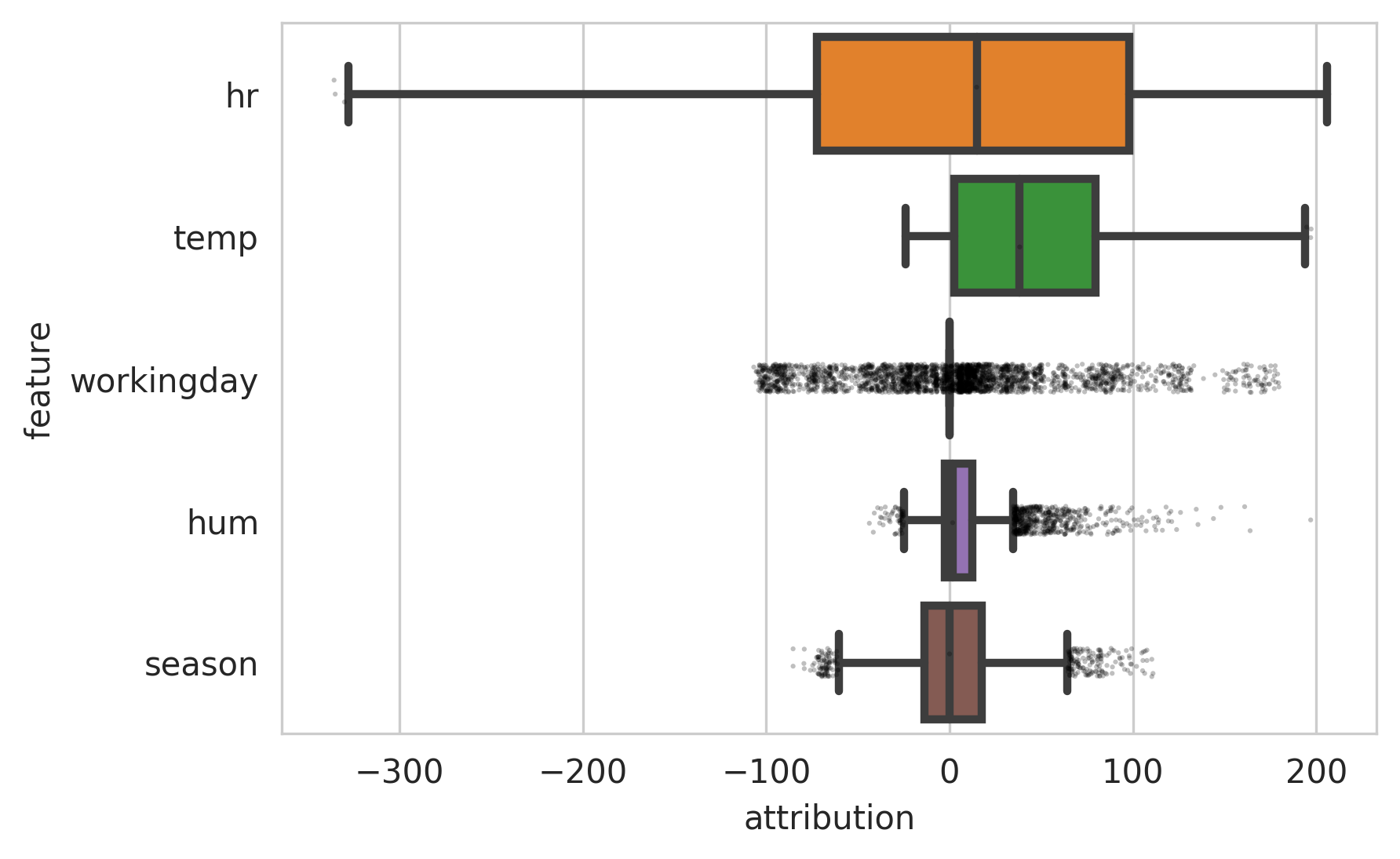}
  \caption{Distribution of single-reference game attributions relative to the data distribution ($\distrinp$) for the Bike Sharing example.}
  \label{fig: bikeshare_boxplot}
\end{figure}

In Section~\ref{subsec:fae}, we discussed that the mean attribution can potentially be a misleading summarization.
Here, we illustrate this using the attributions from the KernelSHAP game $\gamefmarg$ for the Bike Sharing example; see Table~\ref{table:bikeshare_attributions}.
The mean attribution to the feature \textit{hr} is tiny, suggesting that the feature has little impact. However, the distribution of single-reference game attributions (Figure~\ref{fig: bikeshare_boxplot}) reveals a large spread centered close to zero. In fact, we find that by absolute value \textit{hr} receives the largest attribution in over 60\% of the single-reference games. Consequently, only examining the mean of the distribution may be misleading.

\begin{table}[!t]
\centering
\small
    \begin{tabular}{lllllll}
        \toprule
        Game  & Avg. Prediction ($\phi_0$) & hr   & temp & work. & hum & season \\ \midrule
        $\gamefmarg$            & 151             & 3    & 47   & 1          & 7   & 2      \\
        $\gamefjmarg$           & 141             & 6    & 50   & 1          & 9   & 3      \\
        $\gamefunif$            & 128             & 3    & 60   & 3          & 12  & 3      \\ 
        \bottomrule
    \end{tabular}
\caption{Bike Sharing comparison of mean attributions. 95\% CIs ranged from $\pm$0.4 (\textit{hum} in $\distrinp$ and $\distrjm$) to $\pm$2.5 (\textit{hr} in $\distrinp$ and $\distrjm$).}
\label{table:bikeshare_attributions}
\end{table}

\subsubsection{Unquantified uncertainty}\label{sec:case_study_unquantified_uncertainty}
Lack of uncertainty quantification in existing techniques can result in misleading attributions. For instance, taking the mean attribution of 100 randomly-sampled Bike Sharing single-reference games\footnote{The official implementation of KernelSHAP \cite{lundberg2017unified} raises a warning if over 100 references are used.} gives \textit{hr} an attribution of -10 and \textit{workingday} an attribution of 8. Without any sense of uncertainty, we do not know how accurate these (estimated) attributions are. The 95\% confidence intervals (estimated using the method described in Section~\ref{sec: quantify_uncertainty}) show that they are uncertain indeed: the CIs span both positive and negative values.

\subsubsection{Irrelevant references}\label{sec:case_study_unquantified_uncertainty}
In Section~\ref{subsec:fae}, we noted the importance of relevant references (or norms), and how the IME game $\gamefunif$ based on the uniform distribution $\distrunif$ can focus on irrelevant references. We illustrate this on the Adult Income example; see the third row of Table~\ref{table:adult_clusters}. We find that almost all attribution from the $\gamefunif$ game falls on the \textit{capitalgain} feature. 
This is surprising as \textit{capitalgain} is zero for the example being explained, and for over 90\% of examples in the Adult Income dataset. The attributions are an artifact of uniformly sampling reference feature values, which causes nearly all references to have non-zero capital gain (as the probability of sampling an exactly zero capital gain is infinitesimal).

\subsection{Applying the FAE framework}
We now consider how the FAE framework enables more faithful explanations for the three models we study.

\begin{table}[!t]
\centering
\begin{tabular}{llllllll}
\toprule
Game  & Size & Avg. Prediction ($\phi_0$) & rel. & cap. & edu. & mar. & age   \\ \midrule
$\gamefmarg$     & -        & 0.24            & -0.04        & -0.03       & -0.01         & -0.10          & -0.00 \\
$\gamefjmarg$   & -        & 0.19            & -0.02        & -0.03       & -0.01         & -0.08          & 0.01  \\
$\gamefunif$ & -        & 0.82            & 0.01         & -0.79       & 0.02          & -0.03          & 0.04  \\ \midrule
Cluster 1    & 10.2\%       & 0.67            & -0.15        & -0.01       & -0.15         & -0.28          & -0.02 \\
Cluster 2    & 55.3\%       & 0.04            & 0.01         & 0.00        & 0.00          & -0.01          & 0.02  \\
Cluster 3    & 4.4\%        & 0.99            & -0.04        & -0.70       & -0.06         & -0.12          & -0.01 \\
Cluster 4    & 28.0\%       & 0.31            & -0.09        & 0.00        & 0.08          & -0.21          & -0.03 \\
Cluster 5    & 2.1\%        & 0.67            & -0.04        & 0.01        & -0.47         & -0.14          & 0.03  \\ 
\bottomrule
\end{tabular}
\caption{Adult Income comparison of mean attributions from existing game formulations (top) and from clusters obtained from k-means clustering of the single-reference game attributions relative to the input distribution (bottom). 95\% CIs ranged from $\pm$0.0004 (Cluster 2, \textit{relationship}) to $\pm$0.0115 (Cluster 5, \textit{marital-status} and \textit{age}).}
\label{table:adult_clusters}
\end{table}

\begin{table}[!t]
\centering
\small
    \begin{tabular}{lllllll}
        \toprule
        Game  & Avg. Prediction ($\phi_0$) & fico.   & addr. & inc.   & acc.     & dti \\ \midrule
        $\gamefmarg$            & 0.14             & 0.00    & 0.03  & 0.00  & 0.10  & 0.00      \\
        $v_{\inp, \distraccept}$            & 0.05             & 0.02    & 0.04   & 0.02 & 0.11   & 0.03      \\
        \bottomrule
    \end{tabular}
\caption{Lending Club comparison of mean attributions from the game $\gamef$ (relative to the data distribution $\distrinp$) and the game $\gamefgeneric$ (relative to the distribution of accepted applications $\distraccept$).
95\% CIs ranged from $\pm$0.0004 to $\pm$0.0007 for both games.}
\label{table:loan_attributions}
\end{table}

\subsubsection{Formulating contrastive questions}
A key benefit of FAE is that it enables explaining predictions relative to a selected group of references. 
For instance, in the Lending Club model, rather than asking ``Why did our rejected example receive a score of 0.28?'' we ask the contrastive question ``Why did our rejected example receive a score of 0.28 \emph{relative to the examples that were accepted}?''
This is a more apt question, as it explicitly discards irrelevant comparisons to other rejected applications. 
In terms of game formulation, the contrastive approach
amounts to considering single-reference games where the reference is drawn from the distribution of accepted applications (denoted by $\distraccept$) rather than all applications.
The attributions for each of these questions (shown in Table~\ref{table:loan_attributions}) turn out to be quite different. For instance, although number of recently-opened accounts (\textit{acc}) is still the highest-attributed feature, we find that credit score (\textit{fico}), income (\textit{inc}), and debt-to-income ratio (\textit{dti}) receive significantly higher attribution in the contrastive formulation. 
Without formulating the contrastive question, we would be misled into believing that these features are unimportant for the rejection.

\subsubsection{Quantifying uncertainty}\label{sec: quantify_uncertainty}
When summarizing the attribution distribution with the mean,  confidence intervals can be computed using the standard error of the mean (see Section~\ref{sec:ci_theory}). Returning to our Bike Sharing example, with 100 samples, the 95\% confidence intervals for \textit{hr} and \textit{workingday} are -36 to 15, and -1 to 12, respectively. The large CIs caution us that 100 samples are perhaps too few. When using the full test set, the 95\% CIs drop to 0.0 to 5.1 for \textit{hr}, and 0.6 to 2.0 for \textit{workingday}. 

\subsubsection{Summarizing attribution distributions}\label{sec:clustering_case_study}
To obtain a more faithful summarization of the single-reference game attributions, we explore a clustering based approach.
We compute attributions for single reference games relative to points sampled from the the input distribution ($\distrinp$), and then apply $k$-means clustering to these attributions.
The resulting clusters effectively group references that yield similar (contrastive) attributions for the prediction at the explained point.
Consequently,the attribution distribution within each cluster has a small spread, and can be summarized via the mean.

We applied this approach to all three models and obtained promising results, wherein, clustering helps mine distinct attribution patterns.
Table~\ref{table:adult_clusters} (bottom) shows the results for the Adult Income model; results for other models are provided in Appendix~\ref{appendix:case_studies}. Notice that clustering identifies a large group of irrelevant references (cluster 2) which are similar to the explained point, demonstrating low attributions and predictions. Cluster 3 discovers the same pattern that the $\gamefunif$ formulation did: high \textit{capitalgain} causes extremely high scores.
Since over 90\% of points in the dataset have zero \textit{capitalgain}, this pattern is ``washed out in the average''  relative to the entire data distribution $\distrinp$ (as in KernelSHAP); see the first row of Table~\ref{table:adult_clusters}. On the other hand, the IME formulation identifies nothing but this pattern.
Our clustering also helps identify other patterns.
Clusters 1 and 5 show that when compared to references that obtain a high-but-not-extreme score, \textit{marital-status}, \textit{relationship}, and \textit{education-num} are the primary factors accounting for the lower prediction score for the example at hand.

\section{\uppercase{Conclusion}}
We perform an in-depth study of various Shapley-value-based model explanation methods. We find cases where existing methods yield counter-intuitive attributions, and we trace these misleading attributions to the cooperative games formulated by these methods. We propose a generalizing formulation that unifies attribution methods, offers clarity for interpreting each method's attributions, and admits straightforward confidence intervals for attributions.

We propose a conceptual framework for model explanations, called \emph{formulate, approximate, explain} (FAE), which is built on principles from cognitive psychology. 
We advise practitioners to \emph{formulate} contrastive explanation questions that specify the references relative to which a prediction should be explained, for example ``Why did this rejected loan application receive a score of 0.28 \emph{in contrast to the applications that were accepted?}''
By \emph{approximating} the Shapley values of games formulated relative to the chosen references, and
\emph{explaining} the distribution of approximated Shapley values,
we provide a more relevant answer to the explanation question at hand. 

Finally, we conclude that axiomatic guarantees do not inherently guarantee relevant explanations, and that game formulations must be constructed carefully. In summarizing attribution distributions, we caution practitioners to avoid coarse-grained summaries, and to quantify any uncertainty resulting from any approximations used.

\bibliographystyle{splncs04}
\bibliography{references}
\pagebreak
\appendix

\begin{center}
\textbf{\LARGE Appendix}
\end{center}

\section{Shapley Value Axioms}\label{appendix:shapley_axioms}
We briefly summarize the four Shapley value axioms.
\begin{itemize}
\item The \emph{Dummy} axiom requires that if player $i$ has no possible contribution (i.e. $\puregamef(S \cup \Set{i}) = \puregamef(S)$ for all $S \subseteq \players$), then that player receives zero attribution. 
\item The \emph{Symmetry} axiom requires that two players that always have the same contribution receive equal attribution, 
Formally, if $\puregamef(S \cup \Set{i}) = \puregamef(S \cup \Set{j})$ for all $S$ not containing $i$ or $j$ then $\phi_i(v) = \phi_j(v)$. 
\item The \emph{Efficiency} axiom requires that the attributions to all players sum to the total payoff of all players. Formally, $\sum_i \phi_i(\puregamef) = \puregamef(\players)$). 
\item The \emph{Linearity} axiom states that for any payoff function $v$ that is a linear combination of two other payoff functions $u$ and $w$ (i.e. $v(S) = \alpha u(S) + \beta w(S)$), the Shapley values of $v$ equal the corresponding linear combination of the Shapley values of $u$ and $w$ (i.e. $\phi_i(v) = \alpha \phi_i(u) + \beta \phi_i(w)$).
\end{itemize}


\section{Additional Shapley value approximations}\label{appendix:shapley_approx}



\subsubsection{Marginal contribution sampling}
We can express the Shapley value of a player as the expected value of the weighted marginal contribution to a random coalition $S$ sampled uniformly from all possible coalitions excluding that player, rather than an exhaustive weighted sum. A sampling estimator of this expectation is by nature unbiased, so this can be used as an alternative to the permutation estimator in approximating attributions with confidence intervals.
\small
\begin{equation}\label{eq: marginal_sample}
    \phi_i(\puregamef) = \Ex[S]{ \frac{2^{M - 1}}{M} \binom{M - 1}{\Abs{S}}^{-1} \left(\puregamef(S \cup \Set{i}) - \puregamef(S)\right)}
\end{equation}
\normalsize
Equation~\ref{eq: marginal_sample} can be approximated with a Monte Carlo estimate, i.e. by sampling from the random $S$ and averaging the quantity within the expectation. 

\subsubsection{Weighted least squares}
The Shapley values are the solution to a certain weighted least squares optimization problem which was popularized through its use in the KernelSHAP algorithm. For a full explanation, see \url{https://arxiv.org/abs/1903.10464}.
\vskip -0.2in
\small
\begin{equation}\label{eq: wls}
    \pmb{\phi} = \argmin_{\pmb{\phi}} \sum_{S \subseteq \players} \frac{M - 1}{\binom{M}{|S|} |S| (M - |S|)} \left(\puregamef(S) - \sum_{i=1}^{M} \phi_i \right)^2
\end{equation}
\normalsize

The fraction in the left of Equation~\ref{eq: wls} is often referred to as the Shapley kernel
. In practice, an approximate objective function is minimized. The approximate objective is defined as a summation over squared error on a sample of coalitions rather than over squared error on all possible coalitions. Additionally, the ``KernelSHAP trick'' may be employed, wherein sampling is performed according to the Shapley kernel (rather than uniformly), and the least-squares optimization is solved with uniform weights (rather than Shapley kernel weights) to account for the adjusted sampling. 

To the best of our knowledge, there exists no proof that the solution to a subsampled objective function of the form in Equation~\ref{eq: wls} is an estimator (unbiased or otherwise) of the Shapley values. In practice, it does appear that subsampling down to even a small fraction of the total number of possible coalitions (weighted by the Shapley kernel or uniformly) does a good job of estimating the Shapley values for explanation games. Furthermore, approximation errors in such experiments do not yield signs of bias. However, we do note that using the weighted least squares approximation with our confidence interval equation does inherently imply an unproved assumption that it is an unbiased estimator.

\section{Proofs}\label{appendix:proofs}
In what follows, we prove the lemmas from the main paper. The proofs refer to equations and definition from the main paper.
\subsection{Proof of Proposition 1}
From the definitions of $\gamefgeneric$ (Equation 8) and $\gamefsref$ (Equation 9), it follows that $\gamefgeneric(S) = \Ex[\randref \sim \distrref]{\gamefsrandref(S)}$. Thus, the game $\gamefgeneric$ is a linear combination
of games $\{\gamefsref~\vert~\rinp\in\mathcal{X}\}$ with weights defined by the distribution $\distrref$. From the Linearity axiom of Shapley values, it follows that the Shapley values of the game $\gamefgeneric$
must be a corresponding linear combination of the Shapley values of the games $\{\gamefsref~\vert~\rinp\in\mathcal{X}\}$ (with weights defined by the distribution $\distrref$). Thus, 
$\pmb{\phi}(\gamefgeneric) = \Ex[\randref \sim \distrref]{\pmb{\phi}(\gamefsrandref)}$. $\qed$


\subsection{Proof of Proposition 2}
From Proposition 1, we have $\phi_i(\gamefgeneric) = \Ex[\randref \sim \distrref]{\phi_i(\gamefsrandref)}$.
Thus, to prove this lemma, it suffices to show that for any irrelevant feature $i$, the Shapley
value from the game $\gamefsref$ is zero for all references $r\in\mathcal{X}$. That is,
\begin{equation}\label{proof:model_dummy1}
    \forall\rinp\in\mathcal{X}~\phi_i(\gamefsref) = 0
\end{equation}
From the definition of Shapley values (Equation 1), we have:
\begin{equation}\label{proof:model_dummy2}
\phi_i(\gamefsref) = \frac{1}{M} \sum_{S \subseteq \players \setminus \Set{i}} \binom{M - 1}{\Abs{S}}^{-1} \left(\gamefsref(S \cup \Set{i}) - \gamefsref(S)\right)
\end{equation}
Thus, to prove Equation~\ref{proof:model_dummy1} it suffices to show the marginal contribution ($\gamefsref(S \cup \Set{i}) - \gamefsref(S)$) of an irrelevant feature $i$ to any subset of features $S \subseteq \players \setminus \Set{i}$ is always zero.
From the definition of the game $\gamefsref$, we have:
\begin{equation}\label{proof:model_dummy3}
    \gamefsref(S \cup \Set{i}) - \gamefsref(S) = f(\cinp(\inp, \rinp, S \cup \Set{i})) - f(\cinp(\inp, \rinp, S))
\end{equation}
From the definition of composite inputs $\cinp$ (Equation 3), it follows that the inputs $\cinp(\inp, \rinp, S \cup \Set{i})$ and $\cinp(\inp, \rinp, S)$ agree on all features except $i$. Thus, if feature $i$ is irrelevant, $f(\cinp(\inp, \rinp, S \cup \Set{i})) = f(\cinp(\inp, \rinp, S))$, and consequently by Equation~\ref{proof:model_dummy2}, $\gamefsref(S \cup \Set{i}) - \gamefsref(S) = 0$
for all subsets $S \subseteq \players \setminus \Set{i}$.
Combining this with the definition of Shapley values (Equation 1) proves Equation~\ref{proof:model_dummy1}. $\qed$

\section{Reproducibility}\label{appendix:model_details}
For brevity, we omitted from the main paper many of the mundane choices in the design of our toy examples and case studies. To further transparency and reproducibility, we include them here.

\subsection{Fitting models}
For both case studies, we used the LightGBM package configured with default parameters to fit a Gradient Boosted Decision Trees (GBDT) model. 
For the Bike Sharing dataset, we fit on all examples from 2011 while holding out the 2012 examples for testing. We omitted the \textit{atemp} feature, as it is highly correlated to \textit{temp} ($r = 0.98$), and the \textit{instant} feature because the tree-based GBDT model cannot capture its time-series trend. For parsimony, we refitted the model to the top five most important features by cumulative gain (\textit{hr}, \textit{temp}, \textit{workingday}, \textit{hum}, and \textit{season}). This lowered test-set $r^2$ from 0.64 to 0.63.
For the Adult Income dataset, we used the pre-defined train/test split. Again, we refitted the model to the top five features by cumulative gain feature importance (\textit{relationship}, \textit{capitalgain}, \textit{education-num}, \textit{marital-status}, and \textit{age}). This increased test-set misclassification error from 14.73\% to 10.97\%.

\subsection{Selection of points to explain}
For the Bike Share case study, we sampled ten points at random from the test set. We selected one whose prediction was close to the middle of the range observed over the entire test set (predictions ranged approximately from 0 to 600). Specifically, we selected instant 11729 (2012-05-08, 9pm). We examined other points from the same sample of ten to suggest a random but meaningful comparative question. We found another point with comparable \textit{workingday}, \textit{hum}, and \textit{season}: instant 11362. This point caught our eye because it differed only in \textit{hr} (2pm rather than 9pm), and \textit{temp} (0.36 rather than 0.64) but had a much lower prediction.

For the Adult Income case study, we wanted to explain why a point was scored as likely to have low income, a task roughly analogous to that of explaining why an application for credit is rejected by a creditworthiness model in a lending setting. We sampled points at random with scores between 0.01 and 0.1, and chose the 9880th point in the test set due to its strikingly high \textit{education-num} (most of the low-scoring points sampled had lower \textit{education-num}).

For the Lending Club data, we chose an open-source subset of the dataset that has been pre-cleaned to a predictive task on 3-year loans. For the five-feature model, we selected the top five features by cumulative gain feature importance from a model fit to the full set of features.

\subsection{K-means clustering}
We choose $k=5$ arbitrarily, having observed a general tradeoff of conciseness for precision as $k$ increases. In the extremes, $k=1$ maintains the overall attribution distribution, while $k=N$ examines each single-reference game separately.

\section{Case Study Supplemental Material}\label{appendix:case_studies}

\begin{table}[h]
\small
\vskip -0.33in
\centering
\resizebox{0.75\columnwidth}{!}{%
    \begin{tabular}{llllllll}
        \toprule
        Game Formulation & Size  & Avg. Prediction ($\phi_0$) & hr   & temp & work. & hum & season \\ \midrule
        $\gamefmarg$      &  100\% & 151             & 3    & 47   & 1          & 7   & 2      \\
        $\gamefjmarg$      &   100\%  & 141             & 6    & 50   & 1          & 9   & 3      \\
        $\gamefunif$        & 100\%   & 128             & 3    & 60   & 3          & 12  & 3      \\ 
        \midrule
        Cluster 1      & 12.9\%       & 309             & -86  & 14   & -28        & 3   & -1     \\
        Cluster 2      & 27.6\%       & 28              & 140  & 32   & 0          & 9   & 0      \\
        Cluster 3      & 10.5\%       & 375             & -247 & 58   & 16         & 9   & -1     \\
        Cluster 4      & 32.5\%       & 131             & 31   & 38   & 3          & 4   & 2      \\
        Cluster 5      & 16.5\%       & 128             & -57  & 107  & 13         & 9   & 9      \\ 
        \bottomrule
    \end{tabular}%
}
\caption{Bike Sharing comparison of mean attributions. 95\% CIs ranged from $\pm$0.4 (\textit{hum} in $\distrinp$ and $\distrjm$) to $\pm$2.5 (\textit{hr} in $\distrinp$ and $\distrjm$).}
\label{table:bikeshare_clusters}
\vskip -0.2in
\end{table}

\begin{table}[h]
\small
\vskip -0.33in
\centering
\resizebox{0.65\columnwidth}{!}{%
\begin{tabular}{llllllll}
\toprule
Game & Size & Avg. Prediction ($\phi_0$) & rel. & cap. & edu. & mar. & age   \\ 
\midrule
$\gamefmarg$     & 100\%        & 0.24            & -0.04        & -0.03       & -0.01         & -0.10          & -0.00 \\
$\gamefjmarg$   & 100\%        & 0.19            & -0.02        & -0.03       & -0.01         & -0.08          & 0.01  \\
$\gamefunif$ & 100\%        & 0.82            & 0.01         & -0.79       & 0.02          & -0.03          & 0.04  \\ \midrule
Cluster 1    & 10.2\%       & 0.67            & -0.15        & -0.01       & -0.15         & -0.28          & -0.02 \\
Cluster 2    & 55.3\%       & 0.04            & 0.01         & 0.00        & 0.00          & -0.01          & 0.02  \\
Cluster 3    & 4.4\%        & 0.99            & -0.04        & -0.70       & -0.06         & -0.12          & -0.01 \\
Cluster 4    & 28.0\%       & 0.31            & -0.09        & 0.00        & 0.08          & -0.21          & -0.03 \\
Cluster 5    & 2.1\%        & 0.67            & -0.04        & 0.01        & -0.47         & -0.14          & 0.03  \\ 
\bottomrule
\end{tabular}%
}
\caption{Adult Income comparison of mean attributions. 95\% CIs ranged from $\pm$0.0004 (Cluster 2, \textit{relationship}) to $\pm$0.0115 (Cluster 5, \textit{marital-status} and \textit{age}).}
\vskip -0.33in
\end{table}

\begin{table}[h]
\small
\vskip -0.33in
\centering
\resizebox{0.65\columnwidth}{!}{%
    \begin{tabular}{llllllll}
        \toprule
        Game  & Size & Avg. Prediction ($\phi_0$) & fico.   & addr. & inc.   & acc.     & dti \\ 
        \midrule
        $v_{\inp, \distrref}$   &  20\%       & 0.05             & 0.02    & 0.04   & 0.02 & 0.11   & 0.03      \\
        $\gamefmarg$ & 100\% & 0.14             & 0.00    & 0.03  & 0.00  & 0.10  & 0.00      \\
        $\gamefjmarg$   & 100\% & 0.14             & 0.01    & 0.03  & 0.01  & 0.10  & 0.00      \\
        $\gamefunif$ & 100\%        & 0.11  & 0.05         & 0.07       & -0.01          & 0.03          & 0.02  \\ 
        \midrule
        Cluster 1    & 28.5\%       & 0.11            & 0.01        & 0.06       & 0.00         & 0.08 & 0.01       \\
        Cluster 2    & 24.4\%       & 0.10       & 0.01     & 0.00         & 0.01        & 0.11          & 0.04  \\
        Cluster 3    & 15.4\%        & 0.18            & 0.00        & 0.01       & 0.00         & 0.14          & -0.05 \\
        Cluster 4    & 17.6\%       & 0.16            & -0.01        & 0.01        & 0.03          & 0.09          & -0.01 \\
        Cluster 5    & 14.0\%        & 0.22            & -0.01        & 0.05    & -0.02         & 0.08          & -0.06  \\ 
        \bottomrule
    \end{tabular}%
}
\caption{Lending Club comparison of mean attributions. 95\% CIs ranged from $\pm$0.0004 to $\pm$0.0007 for all games.}
\label{table:loan_clusterss}
\end{table}

\end{document}